\pdfoutput=1

\documentclass[11pt]{article}

\usepackage[final]{coling}

\usepackage{comment}
\usepackage{amssymb}
\usepackage{mdframed}
\usepackage{array}
\usepackage{listings} 
\usepackage{times}
\usepackage{latexsym}

\usepackage[T1]{fontenc}

\usepackage[utf8]{inputenc}

\usepackage{microtype}

\usepackage{inconsolata}

\usepackage{graphicx}

\title{A Career Interview Dialogue System using \\Large Language Model-based Dynamic Slot Generation}

\author{
    Ekai Hashimoto\textsuperscript{1},
    Mikio Nakano\textsuperscript{1}\thanks{Also affiliated with C4A Research Institute, Inc., Tokyo, Japan.},  
    Takayoshi Sakurai\textsuperscript{1}, 
    Shun Shiramatsu\textsuperscript{1}, \\
    \textbf{Toshitake Komazaki\textsuperscript{2}}, 
    \textbf{Shiho Tsuchiya\textsuperscript{3}} \\
    \textsuperscript{1}Nagoya Institute of Technology\\
    \textsuperscript{2}Tokyo Healthcare University\\
    \textsuperscript{3}Kitasato University Hospital \\
    \texttt{e.hashimoto.611@stn.nitech.ac.jp}
}

\begin{document}
\maketitle
\begin{abstract}
This study aims to improve the efficiency and quality of career interviews conducted by nursing managers. 
To this end, we have been developing a slot-filling dialogue system that engages in pre-interviews to collect information on staff careers as a preparatory step before the actual interviews. 
Conventional slot-filling-based interview dialogue systems have limitations in the flexibility of information collection because the dialogue progresses based on predefined slot sets. 
We therefore propose a method that leverages large language models (LLMs) to dynamically generate new slots according to the flow of the dialogue, achieving more natural conversations. 
Furthermore, we incorporate abduction into the slot generation process to enable more appropriate and effective slot generation. 
To validate the effectiveness of the proposed method, we conducted experiments using a user simulator. 
The results suggest that the proposed method using abduction is effective in enhancing both information-collecting capabilities and the naturalness of the dialogue. 
\end{abstract}

\section{Introduction}
In hospitals, nursing managers regularly conduct career interviews in order to support the careers of individual staff members. 
However, nursing managers are faced with the time burden of dealing with information that staff are confiding in them for the first time, and of eliciting problems that the staff have. 
Furthermore, interviews are often focused on resignation or career transitions, which imposes a psychological burden on both nursing managers and staff.

In this study, we aim to improve the efficiency and quality of career interviews and provide support to nursing managers through our dialogue system. 
We developed a dialogue system for pre-interviews to collect information on staff careers in preparation for actual interviews and share this information as a report with nursing managers. 
Through the reports, nursing managers will be able to better understand staff concerns and expectations before conducting interviews and provide a more personalized experience. 

As indicated by \citet{Lucas2014ItsOA}, people tend to disclose more information in interactions with systems compared to humans. 
Therefore, we can expect that staff may disclose information to the system that they find difficult to share with nursing managers. 
Additionally, staff will have the opportunity to reassess their careers through interactions with the system before the interview.

As the purpose of our system is to collect information from staff and share it with nursing managers, we designed it to understand the user's speech through slot-filling and to ask questions to fill in the missing slots. 
For slot-filling and question generation, the system utilizes a large language model (LLM). 
Once 80\% of the slots are filled or a predetermined number of dialogue turns have passed, the system ends the conversation and generates a report.

Conventional slot-filling-based interview dialogue systems operate using a predefined set of slots, which presents two problems. 
The first is the limited ability to collect information, as the system can only inquire about information related to the predefined slots. 
The second challenge is the unnaturalness of the dialogue, as it attempts to extract all slot information from every user.

To address these challenges, our method generates slots dynamically during the conversation using an LLM and asks questions based on the newly added slots. 
This is expected to enhance both the information-collecting ability of the dialogue system and the naturalness of the conversation. 
We also propose a method for generating more effective slots by incorporating an abductive process into the slot-generation prompts.

To validate the effectiveness of the proposed method, we conducted comparative experiments utilizing a user simulator to assess the system's ability to collect information in individual staff situations and the naturalness of the dialogues.

\section{Related Work} 
\subsection{Slot-filling Dialogue Systems using LLMs}
Slot-filling-based dialogue management, where knowledge acquired during the conversation is represented by a set of slots and utterances are generated based on that slot set, has long been utilized in dialogue systems \cite{Bobrow1977-BOBGAF}. 
Recently, LLMs have increasingly been used for slot-filling (dialogue state tracking) in dialogue systems \cite{hudecek-dusek-2023-large,siddique_2021,coope-etal-2020-span} to leverage their advanced language understanding capabilities. 
In particular, models like GPT have achieved higher accuracy compared to previous LLM models \cite{sun-etal-2024-speech,feng-etal-2023-towards,heck-etal-2023-chatgpt}.

\subsection{Interview Dialogue Systems} 
Various interview dialogue systems have been developed \cite{devault-21014,4123399,johnston-etal-2013-spoken,kobori-etal-2016-small,zeng-etal-2023-question,b-etal-2020-automatic-edit,10234087,ge-etal-2023-ask}. 
Particularly relevant to our study are the interview systems used in job interviews \cite{published_papers/30815632, su18_interspeech, su19c_interspeech}. 

\citet{published_papers/30815632} developed a dialogue system that acts as an interviewer in job interviews to probe the candidate's motivations. 
Their system asks a base question followed by two follow-up questions generated by the system. 
First, the system evaluates how well the candidate's response has fulfilled the base question, and then based on the results, it presents the prepared follow-up questions.

If the candidate's response to the follow-up question includes specific keywords, the system generates additional follow-up questions using those keywords. 
However, we believe that Inoue et al.’s method lacks the flexibility needed for application in our study. 
Job interview scenarios are relatively constrained, and predefined questions and scenarios can effectively probe. 
In contrast, career interviews require probing across a wide range of scenarios, including individual career goals, workplace experiences, and personal issues. 
Therefore, a static set of questions may be insufficient.

Our proposed method uses LLMs to generate new, highly relevant probing items (slots) and then comes up with follow-up questions to fill the empty slots. 

\subsection{Generating Slots using LLMs}
\citet{Komada} proposed a method that utilizes dynamic slots to maintain dialogue coherence in tabletop role-playing games (TRPGs), where flexible goal-setting is required. 
Their method generates slots filled with information based on the dialogue history, aiming to preserve coherence within the TRPG scenario.

However, applying Komada et al.'s method to our use case presents the challenge of collecting user information. 
Their method focuses primarily on the role of recording the TRPG's progress and does not aim to develop the scenario or collect new information. 
In contrast, career interviews require probing for information that matches the individual circumstances of each staff member. 
Therefore, the system must dynamically generate empty slots suitable for the context of the dialogue and explore them further.

Our proposed method collects user information by generating empty slots through abductive reasoning. 
We believe this will more effectively extract user information and improve the quality of the dialogue.

\section{Proposed Methods}
We adopted slot filling because it is useful for comprehensively gathering the information required to generate a report useful for nursing managers' interviews.

As stated earlier, there are two challenges in adapting conventional slot-filling-based interview dialogue systems to career interviews. 
The first is the limited ability to collect information. 
Slot-filling dialogue systems ask questions to elicit specific information predetermined by a set of slots. 
If the slot sets are appropriately defined, effective information collection becomes possible, but it is unrealistic to prepare an optimal slot sets for every user.

The second challenge is the unnaturalness of the resulting dialogue. 
Slot-filling dialogue systems tend to ask nearly identical questions to all users based on predefined slot sets. 
If slots related to personal issues are included, even users who are not concerned with such issues may be asked these questions. 
This can result in an unnatural dialogue experience for the user.

To address these two challenges, the proposed method utilizes LLMs to generate slots that take the context of the conversation into account. 
Then, the system asks relevant questions based on the newly added slots. 
This approach is expected to improve both the information-collection ability of the dialogue system and the naturalness of the conversation.

\subsection{Generating Slots Using LLMs}\label{sec:Generating Slots Using LLMs}
Slot generation is conducted through prompt engineering with an LLM. 
The prompts include basic instructions, previous slots, previous dialogue history, and the user's recent utterance. 
The prompt for slot generation contains instructions on how to generate slots related to career topics as well as instructions to generate slots that are as useful as possible for report generation.
The full content of all prompts is provided in the Appendix~\ref{ap:prompt}.

\subsection{Incorporating Abductive Process}
During the initial stage of development, we observed that in the absence of clear rules for generation, the slots generated by the system tended to be divergent. 
As a result, there were cases where the generated slots lacked consistency, and in particular, slots unsuitable for career interviews were generated. 
To address this issue, we propose using an abductive approach as a rule for slot generation.

\citet{takanashi} confirmed that in dietary assessments aimed at ensuring optimal nutritional management, supervisors engage in abductive questioning when significant potential risks are reported by patrol members. 
Abduction is a reasoning process formalized as follows \cite{peirce1974collected}: 
\begin{mdframed}[frametitle={Formulation of Abduction}] 
The surprising fact, \textbf{$C$}, is observed;
But if \textbf{$A$} were true, \textbf{$C$} would be a matter of course,
Hence, there is reason to suspect that \textbf{$A$} is true.
\end{mdframed}

We believe Takanashi's findings can be used to identify significant potential risks in a user's career. 
For example, suppose there is a user who is considering resignation due to unhappiness with working night shifts. 
If their utterance indicates ``an interest in management positions'' the surprising fact $C$ is ``interest in management'' and $A$ could be ``the desire for a management position with fewer night shifts''. 
A ``dissatisfaction with night shifts'' slot can then be generated to explore $A$. 
By asking questions to fill this ``dissatisfaction with night shifts'' slot, it is expected that significant potential risks, such as resignation due to dissatisfaction with night shifts, can be identified.

In short, we aim to identify significant potential risks by allowing the LLM to engage in abductive reasoning in the prompt before generating slots. 
This approach is an example of a method where the model performs step-by-step reasoning to arrive at the final output for complex problems, which is known as Chain of Thought (CoT) \cite{10.5555/3600270.3602070}.
CoT methods have been shown to improve both the consistency of reasoning and the accuracy of outputs.
The full content of all prompts is provided in the Appendix~\ref{ap:prompt}.

\section{Implementation}
This section describes our career interview dialogue system that uses the slot generation method explained in the previous section.
Our system works in Japanese.

\begin{figure*}[tbp]
\centering
 \includegraphics[width=0.7\linewidth]{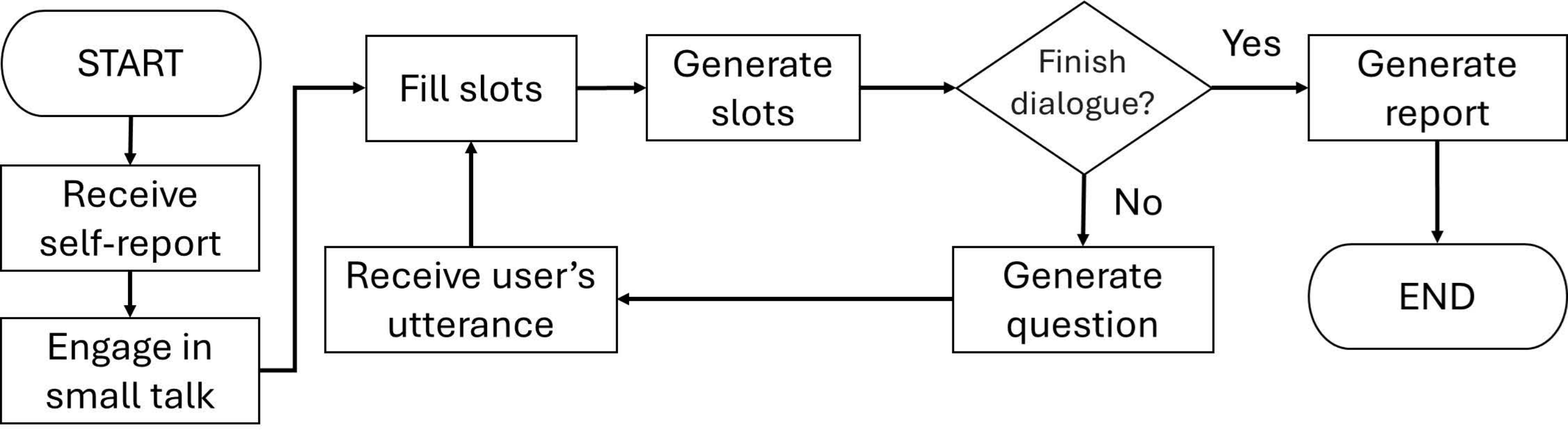}
 \caption{The process flow of our system. (The process for small talk is simplified.)}
 \label{fig:flowchart}
\end{figure*}

The system begins by conducting a dialogue in the small talk phase, where it engages in casual conversation or discussions about concerns with the user. 
After this, the career interview phase starts. Once the interview phase is complete, the system generates a report based on the information collected from the user. 
The user can then review the generated report and select only the information they wish to share with the nursing manager. 
For dialogue management, we utilized the state-transition network-based dialogue management block of DialBB \cite{nakano-sigdial24}. 
The flowchart of the system is shown in Figure~\ref{fig:flowchart}.

\subsection{Self-assessment Questionnaire} 
In a university hospital where one of the authors works, nurses complete a self-assessment questionnaire before the interview. 
This sheet includes both open-ended and multiple-choice questions asking about the nurse's preferences for the next year, training, and career development plans. 
The nursing managers then use this self-assessment questionnaire as the basis for the career interview. 
In our system, users also complete a self-assessment questionnaire before using the system, and the system uses that information during the dialogue.

The format of the self-assessment questionnaire is shown in Table~\ref{tab:Self-assessment Questionnaire}.

\begin{table}[tbp]
\small
\begin{tabular}{p{3cm} p{3cm}}
\hline
\textbf{Question} & \textbf{Options} \\ \hline
\begin{tabular}[t]{@{}p{3cm}@{}}
What are your plans for future career development?\\
(Multiple choice \& Optional free-form description)
\end{tabular} & 
\begin{tabular}[t]{@{}p{4cm}@{}}
- Nursing management \\ 
- Generalist \\ 
- Clinical nurse educator \\ 
- Nurse department faculty \\ 
- Specialized nursing area
\end{tabular} \\ \hline

\begin{tabular}[t]{@{}p{3cm}@{}}
What kind of training would you \\
like to attend? \\
(Select one)
\end{tabular} & 
\begin{tabular}[t]{@{}p{4cm}@{}}
- In-hospital \\ 
- Outside-hospital \\ 
(Specific training name)
\end{tabular} \\ \hline

\begin{tabular}[t]{@{}p{3cm}@{}}
Your preferences for next year \\
(Multiple choice \& Optional free-form description)
\end{tabular} & 
\begin{tabular}[t]{@{}p{4cm}@{}}
- Continue \\ 
- Transfer \\ 
- Resignation \\ 
- Further education
\end{tabular} \\ \hline
\end{tabular}
\caption{Self-assessment questionnaire.}
\label{tab:Self-assessment Questionnaire}
\end{table}

\subsection{Initial Slot Set}
The system starts with eight slots, referred to as the initial slot set, as shown in Table~\ref{tab:initial_slot-set}. 
The initial slot set contains the minimum information required for career interviews as determined by the nursing managers at the university hospital where the experiment was conducted. 
Each slot has a category to provide a broad classification. 
Categorizing the slots allows the system to organize the user's information by category, resulting in a report that is easier for nursing managers to understand.

\begin{table}[tbp]
\small
 \begin{tabular}{l l}
 \hline
 \textbf{Slot name} & \textbf{Category} \\ \hline
 \small Career aspirations for next year & \small Career\\ 
 \small Career development plan & \small Career, Plan\\ 
 \small Future department preferences & \small Career, Preference\\ 
 \small Career-related concerns & \small Career, Concerns\\ 
 \small Training preferences & \small Training, Preference\\ 
 \small Current job duties & \small Job\\ 
 \small Job satisfaction & \small Job, Satisfaction\\ 
 \small Job dissatisfaction & \small Job, Dissatisfaction \\ 
 \hline
 \end{tabular}
 \caption{Initial slot set.}
 \label{tab:initial_slot-set}
\end{table}

\subsection{Dialogue Management in the Interview Phase} 
The system includes a small talk phase before the career interview phase, where the user is encouraged to open up through casual conversation. 
Once career-related topics arise in the small talk, the system transitions to the interview phase. 
In each turn, the system performs slot-filling using an LLM based on the conversation up to that point, and the LLM generates new slots as needed. 
The system also generates questions using the LLM to fill the empty slots. 
The dialogue ends either when a predefined number of turns have passed or when the slot-filling rate exceeds 80\%. 
After the dialogue ends, a report summarizing the slots and conversation history is generated by the LLM.

When using the LLM, the system includes information such as its persona, the content of the self-assessment questionnaire, the slots, and the dialogue history up to that point in the prompt. 
In all modules, the LLM model used is OpenAI's GPT-4o (gpt-4o-2024-05-13) \footnote{https://openai.com/index/hello-gpt-4o/} with a temperature setting of 0.1.

\subsection{Examples of Generated Slots and Questions Among Methods}
\begin{figure}[tbp]
    \hspace{-0.5cm}
    \includegraphics[width=\linewidth]{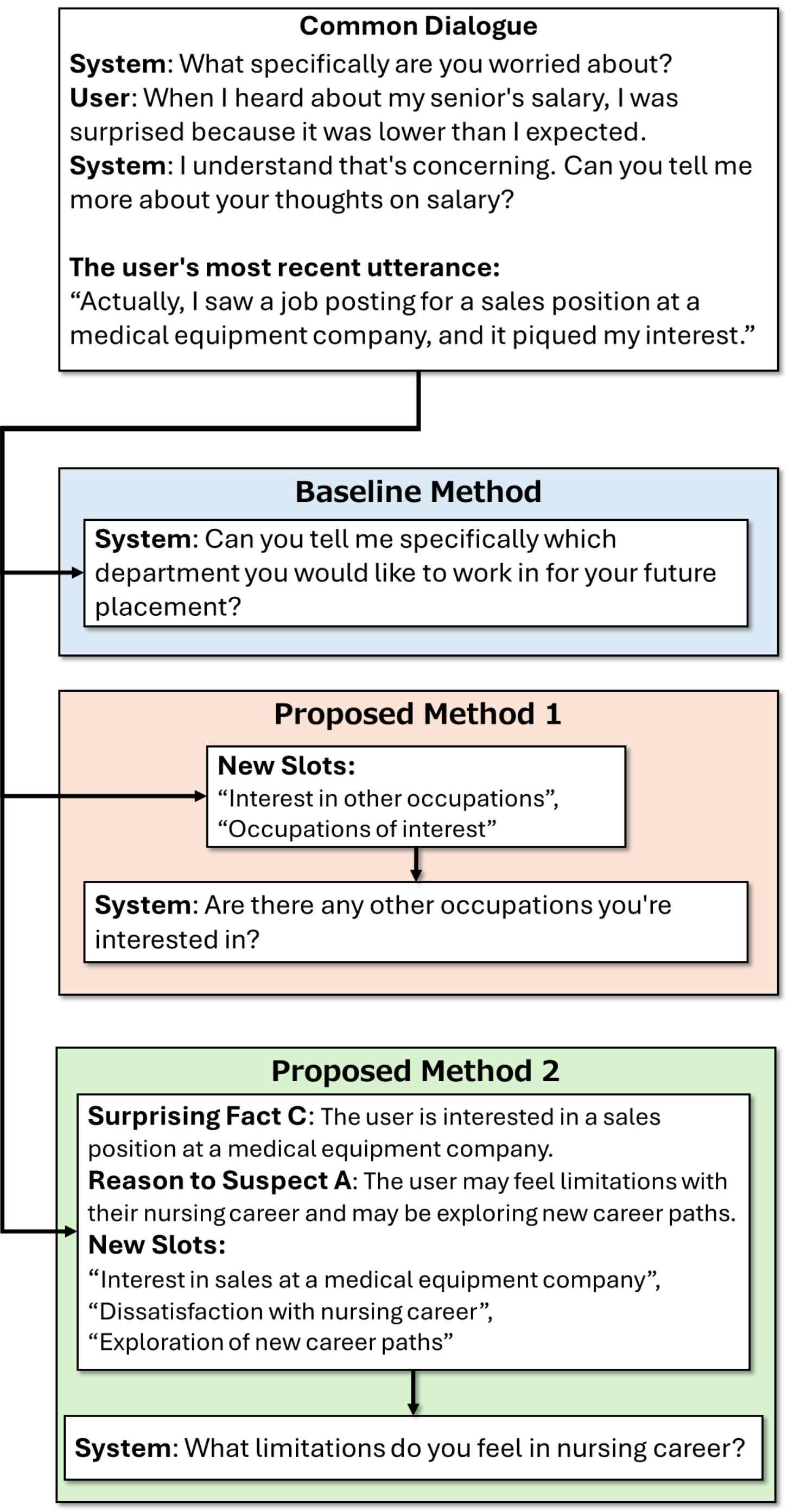}
    \caption{Comparison of generated slots and question among methods.}
    \label{fig:dif_among_methods}
\end{figure}

Figure~\ref{fig:dif_among_methods} shows the system's next question based on the same dialogue history, the user's recent utterance, and slots for each method.
In the Baseline Method, the system asked a question related to the ``future department preferences'' slot, which was included in the initial slot set.
In Proposed Method 1, after receiving the user’s utterance, two new slots were created: ``interest in other occupations'' and ``occupations of interest''. 
The system then asked a question to fill the ``interest in other occupations'' slot. 
In Proposed Method 2, after receiving the user’s utterance, the system first conducted abduction. 
Afterward, three slots were generated to clarify the inferred matters, and the system asked questions to fill the ``dissatisfaction with nursing career'' slot.

\section{Evaluation}
We evaluated and compared the following three methods: Baseline, which uses only the initially set slots without generating any new ones, Proposed Method 1, which generates slots but does not use abductive reasoning, and Proposed Method 2, which performs abductive reasoning before slot generation. 
The slot generation and use of abduction for each method are summarized in Table~\ref{tab:methods_comparison}.

\begin{table}[tbp]
\small
 \centering
 \begin{tabular}{l c c}
 \hline
 \textbf{Method} & \textbf{Slots gen.} & \textbf{Abduction} \\ \hline
 Baseline & $\times$ & $\times$ \\
 Proposed Method 1 & $\checkmark$ & $\times$ \\
 Proposed Method 2 & $\checkmark$ & $\checkmark$ \\ \hline
 \end{tabular}
 \caption{Compared methods.}
 \label{tab:methods_comparison}
\end{table}

The experiment was conducted to verify whether the proposed system can naturally collect information from users through dialogue. 
Specifically, we implemented our system through dialogue with a user simulator based on an LLM and then analyzed the resulting dialogue examples.

Each method was evaluated based on two criteria: ``ability to collect information'' and ``naturalness of dialogues''.

\subsection{Evaluation Items}
\subsubsection{Ability of Collecting Information}
The system's effectiveness as a pre-interview information collection tool for career interviews can be determined by its ability to extract more information within a limited number of turns or time. 
Therefore, considering the persona settings of the user simulator, an average of three items to be extracted were set for each persona. 
We then examined whether these items were mentioned during the dialogue with the proposed system.

Ideally, the evaluation should focus on whether the proposed method can generate appropriate slots. 
However, determining the quality of the generated slots is challenging. Therefore, we adopted a method that directly evaluates whether the necessary information was ultimately obtained.

\subsubsection{Naturalness of Dialogues}
Even with a limited number of turns or time, the system needs to collect information from the user with a natural conversation. 
The content of the system's questions is determined by the slots, which are closely related to the naturalness of the dialogue. 
Therefore, we evaluated how natural the dialogue conducted by each method was.

Our assessment of the naturalness of the dialogue was based on Relevance Theory. 
This theory evaluates the naturalness of dialogues in terms of processing effort and cognitive effects \cite{wilson1995relevance}. 
To investigate the processing effort, we examined the unnaturalness of the system's questioning intentions and the frequency of abrupt topic shifts. 
To investigate cognitive effects, we analyzed how much information the system's questions were able to extract. 
To this end, dialogue examples generated during the experiment were randomly selected and evaluated manually using a 7-point Likert scale on the crowd-sourcing platform Lancers.\footnote{https://www.lancers.jp/}

To examine the processing effort of the system's questions, Evaluation Items 1 and 2 in Table~\ref{tab:Evaluation_items} were set for each dialogue example.
A score of 7 was assigned when the system's questions were considered very natural, and a score of 1 when they were considered very unnatural.

To examine the cognitive effect of the system's questions, Evaluation Items 3 and 4 in Table~\ref{tab:Evaluation_items} were set for each dialogue example. 
A score of 7 was assigned when the system's questions were considered very effective at extracting information, and a score of 1 when they were considered ineffective.

\begin{table}[tbp]
\small
 \begin{tabular}{cp{6.5cm}}
 \hline
 \multicolumn{2}{c}{Effort for processing questions: } \\ \hline
 1: & Did you feel that the system's questions often had unclear intentions? \\ 
 & (1: Strongly agree, 7: Strongly disagree) \\
 2: & Did you feel that there were many abrupt topic shifts in the system's questions? \\ 
 & (1: Strongly agree, 7: Strongly disagree) \\
 \hline
 \multicolumn{2}{c}{Cognitive effect of questions: } \\ \hline
 3: & How effective do you think the system's questions were in extracting information from the nurse?\\ 
 & (1: Very ineffective, 7: Very effective) \\
 4: & Do you think the system's questions were able to extract detailed information from the nurse?\\ 
 & (1: Strongly disagree, 7: Strongly agree) \\
 \hline
 \end{tabular}
 \caption{Evaluation items used to assess the naturalness of dialogues through manual evaluation. (Originally in Japanese)}
 \label{tab:Evaluation_items}
\end{table}

\subsection{Example Dialogues Used for Evaluation}
In this experiment, it is desired to conduct subject experiments under various conditions.
However, such experiments are psychologically burdensome for the participants.
To confirm the proposed method's effectiveness, we conducted a preliminary evaluation through simulations. 
Specifically, we generated dialogues using a user simulator in which each simulated user played a person of a nurse.

Each persona was assigned concerns and other settings, but this information was not provided to the dialogue system, which then interacted with the user simulator.
The user simulator's prompt included instructions to engage in dialogue with the AI and the persona settings. 
The LLM utilized for the user simulator was GPT-4o (gpt-4o-2024-05-13).

A total of 16 personas representing nurses were prepared and self-assessment questionnaires were created for each. 
Each persona's career history and future career plans were defined, followed by a review and evaluation by nursing managers.
Two persona examples are shown in Appendix~\ref{ap:persona_example_endo} and \ref{ap:persona_example_takahashi}.

Through dialogues between the user simulator utilizing these 16 personas and the three methods, a total of 48 dialogue examples were generated. 
These dialogues lasted a minimum of 6 turns and a maximum of 15 turns, with an average of 9.2 turns, including the small talk phase. 

Examples of dialogue are shown in Appendices~\ref{ap:ex_dia_b-m},~\ref{ap:ex_dia_p-m1}, and~\ref{ap:ex_dia_p-m2}.

\subsection{Results}
\subsubsection{Ability of Collecting Information}
For the 16 personas, an average of 3.1 check items per persona was set, resulting in 50 check items in total. 
These items were objectively evaluated by determining whether they were mentioned during the dialogue with the proposed system.


\begin{table}[tbp]
 \centering
 \small
 \begin{tabular}{p{4cm}ccc}
 \hline
 Ave. number & Base & P1 & P2 \\ \hline
 Overall ave. number collected items (upper limit = 3.1) & 2.3 & 2.0 & 2.8 \\ \hline
 Job transfer/resignation ave. number collected items (upper limit = 3.4)& 2.3 &  2.3&  2.9  \\ 
 \hline
 \end{tabular}
 \caption{Average number of collected items in each example dialogue for Overall and Job transfer/resignation-related items. (Base: Baseline 1, P1: Proposed Method 1, P2: Proposed Method 2)}
 \label{tab:collecting_score}
\end{table}

The evaluation of each method's information-collecting ability is shown in Table~\ref{tab:collecting_score}. 
The method that collected the most check items was Proposed Method 2 using abductive reasoning. 
This result demonstrates the effectiveness of slot generation using abductive reasoning.

When limited to 27 check items related to eight personas concerned with job transfer or resignation, Proposed Method 2 collected 23 items (2.9 items on average), while the other two methods collected 18 items (2.3 items on average) each. 
Notably, Proposed Method 1 failed to collect check items from personas with low intent to resign. 

\subsubsection{Naturalness of Dialogues}
To evaluate the naturalness of each method's dialogue, we conducted the same experiment twice using the crowd-sourcing platform Lancers.
In the first experiment, 40 workers rated the dialogues' naturalness, resulting in an effect size of approximately 0.40, with a calculated valid sample size of about 90. 
We then collected additional 48 workers, bringing the total to 88 workers for the analysis.
The dialogue examples used for evaluation were selected randomly from the 48 dialogue examples, with three dialogues (one per method, with the same persona) being evaluated. 
The differences in the evaluation results based on the dialogue management methods were analyzed using Friedman's test and the Wilcoxon signed-rank sum test. 
The results are shown in Table~\ref{tab:naturalness_score}.

\begin{table*}[tbp]
\centering
\small
\begin{tabular}{cc|ccc|>{\raggedleft\arraybackslash}m{1.5cm}rrr}
\hline
\multicolumn{2}{c|}{Question} & \multicolumn{3}{c|}{$mean\:\:{(S.D.) }$} & \multicolumn{4}{c}{p-value} \\
\multicolumn{2}{c|}{items} & Base & P1 & P2 & Base vs. P1 vs. P2 & Base vs. P1 & Base vs. P2 & P1 vs. P2 \\
\hline
Processing&1 & $4.90\:\:{(1.45) }$ & $4.94\:\:{(1.46) }$ & $5.23\:\:{(1.29) }$ & $\dag$ 0.071 & 0.734 & 0.058 & 0.040 \\
effort&2 & $4.45\:\:{(1.41) }$ & $4.33\:\:{(1.55) }$ & $5.06\:\:{(1.48) }$ & $**$0.008 & 0.508 & $\star$ 0.003 & $\star$p < 0.001 \\
\hline
Cognitive &3 & $4.65\:\:{(1.45) }$ & $4.58\:\:{(1.59) }$ & $5.01\:\:{(1.37) }$ & $\dag$ 0.087 & 0.641 & 0.080 & 0.010 \\
effect &4 & $4.55\:\:{(1.39) }$ & $4.58\:\:{(1.37) }$ & $5.14\:\:{(1.31) }$ & $*$0.019 & 0.866 & $\star$0.004 & $\star$0.004\\
\hline
\multicolumn{9}{r}{$\dag$ p<0.1, $*$ p<0.05, $**$ p<0.01: \textit{Friedman's test}}\\
\multicolumn{9}{r}{$\star$ p<0.017: \textit{Wilcoxon signed-rank test with Bonferroni correction for multiple comparisons}}
\end{tabular}
\caption{Results of the evaluation of the naturalness of dialogues (n = 88). (Base: Baseline, P1: Proposed Method 1, P2: Proposed Method 2)}
\label{tab:naturalness_score}
\end{table*}

After conducting Friedman's test across the three methods, Wilcoxon signed-rank tests were performed for all pairs. 
Pairwise comparisons were also adjusted using the Bonferroni method for multiple comparisons.

As shown in Table~\ref{tab:naturalness_score}, Proposed Method 2 achieved the highest average scores in all evaluation items. 
Additionally, it exhibited lower standard deviations for most evaluation items.

\subsubsection{Performance of the Slot-generating Module}
We calculated the average number of generated slots for each user's response based on the generated dialogue examples.
The results are shown in Table~\ref{tab:slot_gen_score}. 
These evaluations confirmed that the slot-generation module's performance, regarding the number of generated slots, meets a certain standard.
\begin{table}[tbp]
 \centering
 \small
 \begin{tabular}{cc}
 \hline
 Methods & {$mean\:\:{(S.D.)}$} in a turn \\ \hline
 P1 &  $2.38\:\:{(1.80) }$ \\
 P2 &  $3.78\:\:{(1.26) }$  \\ 
 \hline
 \end{tabular}
 \caption{Average number of each method's generated slots. (P1: Proposed Method 1, P2: Proposed Method 2)}
 \label{tab:slot_gen_score}
\end{table}

\subsubsection{Performance of the Slot-filling Module}
This evaluation is based on the assumption that the system's language understanding performance, specifically the accuracy of slot filling, is relatively high. 
So, we randomly selected 66 user turns from the generated dialogue examples and calculated the F1 score.
The F1 score of the slot-filling module is shown in Table~\ref{tab:f1_score}. 
The performance of the slot-filling module was confirmed to meet a certain standard through these evaluations.

\begin{table}[tbp] 
\small
    \centering
    \begin{tabular}{ll} \hline
    Items & Score \\ \hline
    Precision & $0.825$  $(85/103)$ \\ 
    Recall & $0.842$  $(85/101)$\\
    F1 & $0.833$\\ \hline
    \end{tabular} 
    \caption{Performance of the slot-filling module.} 
    \label{tab:f1_score} 
\end{table}

\subsubsection{Performance of the User Simulator}
If the user simulator makes unnatural utterances, the entire dialogue will become unnatural regardless of the system's utterances, making this evaluation meaningless. 
Therefore, in addition to the experiment evaluating the naturalness of the dialogue, we added a phase to measure the performance of the user simulator. 
We also asked the crowd-workers who evaluated the naturalness of the example dialogues to asses the user utterances in them using a 7-point Likert scale.
The questions and results utilized for this evaluation are shown in Table~\ref{tab:persona_result}. 
From these results, the performance of the user simulator was confirmed to meet a certain standard.

\begin{table}[!tbp] 
\small
\centering 
\begin{tabular}{p{4.5cm}c} \hline
Questions & $mean\:\:{(S.D.)}$\\ 
\hline
Did the nurse's statements align with the persona settings? (1: Strongly disagree, 7: Strongly agree) & $5.04\:\:{(1.31)}$ \\ \hline
Were the nurse's statements natural? (1: Strongly disagree, 7: Strongly agree) & $5.13\:\:{(1.36)}$ \\ 
\hline
\end{tabular} 
\caption{Performance of the user simulator.} 
\label{tab:persona_result} 
\end{table}

\section{Discussion}\label{sec:Discussion}
\subsection{Ability of Collecting Information}\label{sec:discuss_info}
As shown in Table~\ref{tab:collecting_score}, Proposed Method 2 collected the most check items. 
It was able to infer the reason $A$ whenever the user made a statement suggesting they might leave their current job. 
To clarify $A$, the system generated a slot related to $A$ and posed a corresponding question. 
This process enabled the system to collect more latent risks and related information (reasons and circumstances) from the users.

The reason Proposed Method 1 collected fewer items than the Baseline is likely due to its inability to control which items should be probed further. 
As shown in Table~\ref{tab:collecting_score}, although the score for personas with high intent to leave was the same as the Baseline, Proposed Method 1 collected five fewer items (0.3 items on average) overall. 
One common example where Proposed Method 1 failed to collect check items was when it focused too narrowly on a single issue (particularly current job duties). 
As a result of probing too deeply into a narrow topic, Proposed Method 1 was unable to collect a wider range of check items. 
The LLM determines which items to inquire about based on the prompt. 
When the conversation shifts to a potentially concerning topic, the system keeps asking about that topic. 
This is likely why Proposed Method 1 collected the fewest items.

\subsection{Naturalness of Dialogues}
\subsubsection{Processing Effort of Questions}
Although there was no statistically significant difference in Evaluation Item 1 in Table~\ref{tab:Evaluation_items}, Proposed Method 2 produced the best results. 
Proposed Method 2 dynamically generated questions based on abductive reasoning following the user's responses, asking questions to test the hypothesis. 
Since the user may not fully understand the hypothesis or the reasoning process, this could have led to situations where the questions seemed unclear in intent. 
As a potential solution, presenting the reasoning process behind the hypotheses to the user might reduce the number of questions with unclear intentions. 

In Evaluation Item 2, Proposed Method 2 showed statistically significant differences compared to the other methods and achieved relatively high results. 
As with the reasoning presented in Evaluation Item 1, Proposed Method 2 used an abductive reasoning approach for generating questions.
This approach allowed for a natural flow of conversation and reduced the occurrence of sudden topic shifts. 
However, Proposed Method 2 did not achieve sufficiently high results.
Proposed Method 2 generated different hypotheses in response to the user's responses, which could lead to a variety of topics being explored. 
As a result, users may have encountered unexpected questions, leading evaluators to feel that the system was abruptly changing topics.

In Evaluation Item 2, the Baseline Method and Proposed Method 1 achieved lower results compared to Proposed Method 2.
In dialogues created using Proposed Method 1 during this experiment, topic shifts occurred when the LLM determined that further exploration was not feasible. 
This was particularly evident when the slots for deepening the discussion were not appropriately generated, leading to abrupt topic changes.
Compared to Proposed Method 2, the slots generated by Proposed Method 1 for deeper exploration were less suitable, which likely contributed to its lower evaluation.
In contrast, the Baseline Method switched topics whenever one of the initial slot set in the predefined set was completed. Due to the weak relationships between the slots in the initial set, evaluators likely perceived the topic shifts to new slots as abrupt.

\subsubsection{Cognitive Effect of Questions}
Although there was no statistically significant difference in Evaluation Item 3, Proposed Method 2 produced the best results. 
Proposed Method 2 dynamically generated slots and hypotheses using abductive reasoning based on user responses, guiding the dialogue through hypothesis testing.
This strategy resulted in a more natural conversational flow and helped prevent topic drift, even when handling multiple slots. 
As a result, it is believed that the method was able to extract the necessary information more effectively.

On the other hand, Proposed Method 1 had the lowest score and the largest standard deviation. 
This can be attributed to its approach of generating ``slots suitable for career counseling'' to guide the dialogue. 
However, it often focused excessively on specific topics (e.g., current tasks) or exhibited weak associations between slots, leading to scattered questions. 
Consequently, it was evaluated as less effective in collecting information.

In Evaluation Item 4, Proposed Method 2 showed statistically significant differences compared to the other methods and achieved relatively high results. 
Statistically significant differences were observed between the Baseline Method and the proposed methods.
Proposed Method 2, which employs abductive reasoning for probing, asked deeper and more detailed questions, improving the overall quality of questions throughout the dialogue.

\section{Concluding Remarks}
In this study, we developed a slot-filling dialogue system that generates slots to support the preparation for career interviews through the collection of user information.

We conducted an experiment utilizing the user simulator to verify the effectiveness of the proposed methods, evaluating its ability to collect information and the naturalness of the dialogue. 

The results suggest that slot generation using abductive reasoning contributes to more effective information collection from users. 
However, we also observed that focusing on information collecting led to a loss of naturalness in the dialogue, with some questions lacking clear intent.
We plan to conduct experiments with actual nurses and further improve the effectiveness of slot generation.

Since the experiments in this study were conducted with the user simulator, it remains unclear how the system will perform in real interactions with nurses. 

\section{Limitations}
This study has several limitations. 
The most significant is the issue concerning the accuracy of slot generation. 
In our method, slot generation is performed using an LLM based on abductive reasoning, but there is a risk of generating inappropriate slots due to excessive inference or misinterpretation. 
LLMs can sometimes generate slots that do not fit the context or are based on non-existent information, leading to unnatural questions that disrupt the flow of the dialogue. 
Preventing such mis-generation and improving the accuracy of the generated slots will be a key challenge in future research.

Excessive slot generation can result in overly long prompts, potentially compromising the naturalness of speech and the consistency of the dialogue. 
The core issue lies in retaining too many slots, which can lead to inappropriate responses. 
To address this, first, we aim to filter out less relevant slots, thereby shortening the prompt. 
Second, we are considering a design where users can stop the dialogue halfway through if inappropriate responses are generated, reducing the psychological burden.

The second limitation is related to the question generation module. 
If this module creates questions based on inappropriate slots, it may ask ineffective questions, negatively impacting information collection and diminishing the naturalness of the dialogue. 
Therefore, further optimization of the question generation module is necessary.

The final limitation is the use of the user simulator for evaluation. 
Although the results were promising, our system has not yet been verified using interactions with actual nurses. 
Real users may exhibit unexpected statements or emotional reactions, which could alter the system's performance and behavior. 
Thus, evaluating the system with real users is essential for future research.

Considering these limitations, future studies should focus on refining the slot generation algorithm, improving the question generation module, and conducting evaluations with real users.

\section{Ethical Considerations}

When this system is deployed, users will share their personal information. 
However, the dialogue history will not be shared with the nursing manager. 
Before sharing the report to the nursing manager, the users have the opportunity to review and edit its contents. 
Therefore, there are no privacy concerns.

Potential risks include the possibility of the system being used for training purposes and the risk of OpenAI or the system's servers being hacked. 
However, this system does not require users to input their names. 
By linking users to their employee ID numbers, the risk of external identification can be kept low.
Additionally, since users can edit the report content, if logs are not stored, privacy can be effectively maintained.
 

\section*{Acknowledgment}
This study is partially supported by JST CREST (JPMJCR20D1), NEDO (JPNP20006) and JSPS KAKENHI(24K03052, 22K12325).

\bibliography{custom,anthology}

\clearpage

\appendix
\section{Prompts Used in the System} 
\label{ap:prompt}
%
Figure~\ref{fig:prompt_prop1} shows the prompt for Proposed Method 1. 
Figure~\ref{fig:prompt_prop2_1st} and \ref{fig:prompt_prop2_2nd} show the prompt for Proposed Method 2. 
Figure~\ref{fig:prompt_slot-fill} shows the prompt for slot filling. 
Figure~\ref{fig:prompt_gen_q} shows the prompt for generating questions. 
Figure~\ref{fig:prompt_user_sim} shows the prompt for the user simulator. 

The texts enclosed in < and > in Figures~\ref{fig:prompt_prop1}, \ref{fig:prompt_prop2_2nd}, \ref{fig:prompt_slot-fill}, ~\ref{fig:prompt_gen_q} and \ref{fig:prompt_user_sim} represent placeholders. 
For instance, ``<Dialogue History>'' denotes the section of the prompt where the actual user-system interaction up to the current turn is inserted.
\begin{figure*}[tbp]
\small
\begin{verbatim}
Instructions:
You are a career consultation support system for nurses. The main function of this system is to collect 
and organize information related to a nurse's career. In career consultations, specific "information 
slots" are used to collect and record information about the nurse's career. Your role is to create 
appropriate new slots based on the information obtained from the conversation history with the nurse or 
from their self-assessment questionnaires. These new slots will be used in questions.

How to Create a New Slot:
Analysis of Conversation History: Analyze the goals, interests, and concerns related to both career and 
personal matters as expressed by the nurse during the consultation. Categorization of Information: 
Categorize the analyzed information into groups, such as "career goals," "specialized skills," "career 
concerns," or "personal concerns."

Defining Slots:
Based on each category, create new slots. It is important to include a clear name and detailed 
description for each slot. Assignment of Information: Appropriately assign relevant information to the 
corresponding slots.

Things You Must Not Do:
Do not generate slots that are identical to existing ones.

Slot Description:
In the career consultation support system for nurses, slots are used to organize and systematically 
record various types of information and data related to career development. Each slot corresponds to a 
specific topic or question and reflects concrete aspects related to the nurse's career. These may 
include career goals, specialized skills, intentions for further education, interests in training 
programs, as well as non-career-related information.

Instructions for Output:
Always output in JSON format. Do not include previous slots in the output. Do not include the content in 
the "value" field for new slots. Ensure the "value" is always set to null

Output Example (multiple slots are acceptable) :
{ # A maximum of 5 new slots can be generated
    "XXX": { # Slot name
        "category": "xxx,zzz", # Slot category
        "value": null # Always set to null
    },
    "YYY": { # Slot name
        "category": "yyy, zzz", # Slot category
        "value": null # Always set to null
    }
}

Dialogue History:
< Dialogue History >

Current Slots:
< Current Slots >
\end{verbatim}
\caption{Prompt used for Proposed Method 1. (Originally in Japanese)}
\label{fig:prompt_prop1}
\end{figure*}
\begin{figure*}[tbp]
\small
\begin{verbatim} 
Instructions:
You are a career consultation support system for nurses. The main function of this system is to collect 
and organize information related to a nurse's career. In career consultations, specific "information 
slots" are used to collect and record information about the nurse's career. Your role is to create 
appropriate new slots based on the information obtained from the conversation history with the nurse or 
from their self-assessment questionnaires. These new slots will be used in questions.

How to Create a New Slot:
If necessary, use the following "abduction" approach when creating a new slot.

Abduction:
"A surprising fact C is observed. However, if A were true, C would naturally follow; therefore, there 
is reason to suspect A is true."
Surprising fact C: This refers to particularly surprising facts (such as the intention to resign) 
regarding the nurse's career. These surprising facts greatly exceed normal expectations or predictions.
Reason to suspect A: This refers to problems, reasons, or grounds that explain the observed surprising 
fact C (such as being too busy at work)  experienced by the user. It also refers to any inferred 
information or situation based on that. You should consider the surprising fact C and inferred matter A 
in context.

Specific Steps:
Analysis of Conversation History: Use abduction only when surprising facts C related to career or 
personal goals, interests, and concerns are mentioned by the nurse during the consultation. 
Categorization of Information: Categorize the analyzed information into groups such as "career goals," 
"specialized skills," "career concerns," and "personal concerns."

Defining Slots: 
Based on each category, create new slots. It's important to include a clear name and a 
detailed description for each slot. Assignment of Information: Assign relevant information to the 
appropriate slots.

Creating Slots:
Ensure diversity in categories when creating slots. If inferred matter A exists, create a new slot 
based on C. If there is no inferred matter A, create a new slot based on the conversation history or 
the user's current slots. Do not create slots that are identical or similar to existing ones.
\end{verbatim}
\caption{Prompt used for Proposed Method 2. (first half) (Originally in Japanese)}
\label{fig:prompt_prop2_1st}
\end{figure*}
\begin{figure*}[tbp]
\small
\begin{verbatim}
Slot Description:
In the career consultation support system for nurses, slots are used to organize and systematically 
record various types of information and data related to career development. Each slot corresponds to a 
specific topic or question and reflects concrete aspects related to the nurse's career. These may 
include career goals, specialized skills, intentions for further education, interests in training 
programs, as well as non-career-related information.

Things You Must Not Do:
Do not generate slots that are identical to existing ones. Do not repeat the same abduction as before 
by referencing the abduction history. However, it is acceptable if the inferred matter differs.

Slot Description:
In the career consultation support system for nurses, slots are used to organize and systematically 
record various information and data related to career development. Each slot corresponds to a specific 
topic or question and reflects concrete aspects related to the nurse's career. It may also include 
non-career-related information.

Instructions for Output:
Always output in JSON format. Do not include previous slots in the output. Do not include the content 
in the "value" field for new slots. Ensure the "value" is always set to null.

Output Example (multiple slots are acceptable) :
{ 
  "Surprising Fact C": "", # Content of C. If none, set to null  
  "Reason to Suspect A": "", # Content of A. If none, set to null  
  "New Slot": { # A maximum of 5 new slots can be generated
        "XXX": { # Slot name
            "category": "xxx,zzz", # Slot category
            "value": null # Always set to null
        },
        "YYY": { # Slot name
            "category": "yyy, zzz", # Slot category
            "value": null # Always set to null
        }
    }
}

Dialogue History:
< Dialogue History >

Current Slots:
< Current Slots >

Abduction History:
< Abduction History >
\end{verbatim}
\caption{Prompt used for Proposed Method 2. (second half) (Originally in Japanese)}
\label{fig:prompt_prop2_2nd}
\end{figure*}
\begin{figure*}[tbp]
\small
\begin{verbatim} 
Instructions:
You are a career consultation support system for nurses. The main function of this system is to collect 
and organize information related to a nurse's career. In career consultations, specific "information 
slots" are used to collect and record information about the nurse's career. Your role is to identify 
the appropriate slots based on the conversation history with the nurse or their self-assessment 
questionnaires and fill them out with specific details. "Specific" means using the words and terms 
expressed by the user in their responses. You can also add or update information in slots that have 
already been filled.

How to Fill Slots:
Based on Consultation: Analyze the conversation history from the consultation and identify the nurse's 
stated career and personal goals, interests, and concerns. Assign the relevant information or user 
statements to the corresponding slots.

Slot Description:
In the nurse career consultation support system, slots are used to organize and systematically record 
various information and data related to career development. Each slot corresponds to a specific topic 
or question and reflects concrete aspects related to the nurse's career. For example, slots can include 
career goals, specialized skills, intentions for further education, interest in training programs, as 
well as non-career-related information.

Output Instructions:
Always output in JSON format. Do not delete any slots. Output all the slots. For example, if there is 
a slot named "Career Hope for Next Year" and the user answers "I want to continue working," add the word 
"continue" to that slot. When there are similar slots, do not create new ones; instead, add the 
information to the existing slot. Do not create new slots; always fill in the information in existing  
slots.

Output Example:
{
    "XXX":{ # Slot name
        "category" : "aaa,bbb", # Slot category
          "value" : "ccc", # Fill slot
    },
    "YYY": { #  Slot name
        "category" : "ddd,eee", # Slot category
        "value" : "fff", # Fill in the slot content; the initial value is null
    }
}

Dialogue History:
< Dialogue History >

Current Slots:
< Current Slots >
\end{verbatim}
\caption{Prompt used for slot-filling. (Originally in Japanese)}
\label{fig:prompt_slot-fill}
\end{figure*}
\begin{figure*}[tbp]
\small
\begin{verbatim}
Situation:
You are an experienced nurse. You are about to listen to a junior nurse's concerns regarding their 
career and personal life.

Task: 
In order to understand the user's current situation, ask questions based on the provided slots to fill 
them in. The questions should be designed to draw out the user's thoughts. Keep the question within 100 
characters and limit it to one question. If there are slots with similar categories, aim for an 
efficient question. Follow a one-question, one-answer format, and crossing slots with your questions is 
only allowed if the questions are logically related. Be clear in your questions, including subject and 
object, and avoid abstract questions. Never repeat the same question.

Your Persona
Name: Keiko Naasu
Age: 34
Personality Traits:
Kind and compassionate, always treating both patients and colleagues with care. Has a wealth of 
professional experience, making sound judgments in various situations. She is also patient in teaching 
new nurses. Calm and composed, able to handle emergencies with poise. Values teamwork and has excellent 
communication skills.
Background:
Over 10 years of nursing experience with work in various departments. Actively participates in the 
hospital's mentoring system, supporting the growth of junior nurses.
Speaking Style:
Speaks in a friendly and supportive manner toward junior nurses. Uses casual speech.
Output Instructions:
The output must always be in JSON format.
Output Example
{
  "Target Slot S": { # Slot name
    "XXX": { # Slot name
      "category": "xxx,zzz", # Slot category
      "value": null # Always set to null
    },
    "YYY": { # Slot name
      "category": "yyy, zzz", # Slot category
      "value": null # Always set to null
    }
  },
  "Question": "" # Question
}

Dialogue History:
< Dialogue History >

Current Slots:
< Current Slots >
\end{verbatim}
\caption{Prompt used for generating question. (Originally in Japanese)}
\label{fig:prompt_gen_q}
\end{figure*}

\begin{figure*}[tbp]
\small
\begin{verbatim}
Task Description:
- You are a user of the dialogue system, consulting the AI about your career.
- Generate your next utterance in 50 characters or less.
- Even if asked about something unrelated to your persona, respond based on your persona.
- Do not repeat the same utterance multiple times.
- Be mindful of the previous context when responding.
- Do not include your name or "User," just generate the utterance.
- You cannot ask questions.

Your Persona:
< Each Persona's Setting >

Situation: 
- The AI is a senior nurse.
- The AI is someone reliable.
- The AI is someone you can consult with.

Dialogue History:
< Dialogue History >
\end{verbatim}
\caption{Prompt used for the user simulator. (Originally in Japanese)}
\label{fig:prompt_user_sim}
\end{figure*}
\clearpage
\section{An Example of Persona (Aoi Endo) and Dialogues }

\subsection{An Example of Persona Setting: Aoi Endo (Originally in Japanese) }\label{ap:persona_example_endo}
Table~\ref{tab:persona_example_endo} shows the persona setting of Aoi Endo.

\begin{table}[tbp]
\centering
\begin{tabular}{|>{\centering\arraybackslash}p{2cm}|p{4cm}|}
\hline
\textbf{Attribute}           & \textbf{Details}                                                                 \\ \hline
\small \textbf{Name}                & Aoi Endo                                                                          \\ \hline
\small \textbf{Age}                 & 30 years old                                                                                \\ \hline
\small \textbf{Hometown}            & Niigata Prefecture                                                                \\ \hline
\small \textbf{Gender}              & Female                                                                            \\ \hline
\small \textbf{Personality}         & Sincere, cooperative                                                              \\ \hline
\small \textbf{Past Career}         & A nurse working at a university hospital, in her 8th year as a nurse, and has been working there throughout her career. \\ \hline
\small  \textbf{Current Career} & Works in internal medicine and serves as a deputy leader in the team.\\ \hline
\small  \textbf{Future Aspirations}  & Aims to advance into a nursing management position. \\ \hline
\small \textbf{Thoughts} & Satisfied with good relationships at her current workplace but feels dissatisfied with the lack of promotion opportunities. Wishes to continue working at the same hospital. \\ \hline
\small  \textbf{Other Details} & Her hobby is handicrafts. She has one child, and her parents live nearby and provide support for childcare. \\ \hline
\small  \textbf{Check Items} & Intentions toward nursing management positions, Dissatisfaction with promotion, Desire for continuity. \\ \hline
\end{tabular}
\caption{Persona: Aoi Endo.} 
\label{tab:persona_example_endo}
\end{table}

\subsection{An Example Dialogue using Baseline Method with Aoi Endo  (Originally in Japanese) }\label{ap:ex_dia_b-m}
\begin{quote}
\textbf{System}: Have you been busy lately? \\
\textbf{Aoi Endo }: Yes, I'm busy, but I feel it's rewarding. \\
\textbf{System}: Despite being busy, what kind of support do you think would make it easier to work? \\
\textbf{Aoi Endo }: I'm dissatisfied with the lack of promotion opportunities. How can this be improved? \\
\textbf{System}: Regarding the lack of promotion opportunities, what specific support or systems do you think would help improve this situation? \\
\textbf{Aoi Endo }: It would be helpful to have training and support for obtaining qualifications for promotion. \\
\textbf{System}: Regarding the support for training and obtaining qualifications for promotion, what kind of content or format do you think would be ideal? \\
\textbf{Aoi Endo }: Online courses for obtaining qualifications and regular training sessions would be ideal. \\
\textbf{System}: For the online courses and regular training sessions, what specific content or themes do you think would be useful? \\
\textbf{Aoi Endo }: Leadership training for management positions and updates on the latest medical knowledge. \\
\textbf{System}: Regarding the leadership training for management positions and the latest updates on medical knowledge, what specific content or themes do you think would be useful? \\
\textbf{Aoi Endo }: For leadership training, effective team management and communication skills are important. \\
\textbf{System}: Can you tell me more specifically about the team management and communication skills you'd like to learn in the leadership training? \\
\textbf{Aoi Endo }: In terms of team management, effective task delegation and maintaining motivation are important. \\
\textbf{System}: That's all for today!
\end{quote}
\subsection{An Example Dialogue using Proposed Method 1 with Aoi Endo  (Originally in Japanese) }\label{ap:ex_dia_p-m1}
\begin{quote}
\textbf{System}: Have you been busy lately?  \\
\textbf{Aoi Endo}: Yes, I'm busy, but I feel it's rewarding.  \\
\textbf{System}: I see, it's wonderful that you're feeling rewarded! By the way, can you tell me about your current job responsibilities?  \\
\textbf{Aoi Endo}: I'm supporting the team as a deputy leader in the internal medicine department.  \\
\textbf{System}: As a deputy leader, can you tell me specifically what kind of tasks you're responsible for?  \\
\textbf{Aoi Endo}: I'm in charge of coordinating the team, training new staff, and creating care plans for patients.  \\
\textbf{System}: I see, you're handling a variety of roles. Can you tell me more about your specific career goals?  \\
\textbf{Aoi Endo}: I want to move into a nursing management position.  \\
\textbf{System}: So, you're aiming for a nursing management position. Can you share your career development plan in more detail?  \\
\textbf{Aoi Endo}: Since there are few promotion opportunities, I'm also considering other hospitals.  \\
\textbf{System}: You're interested in other hospitals, but what kind of hospitals or environments are you specifically considering?  \\
\textbf{Aoi Endo}: I'm conflicted because the relationships at my current workplace are good.  \\
\textbf{System}: It's great that the relationships are good! Specifically, what aspects of the relationships do you feel are good?  \\
\textbf{Aoi Endo}: My colleagues are cooperative, and it's easy to exchange opinions.  \\
\textbf{System}: You mentioned there are few promotion opportunities. What kind of support do you think would increase the chances of promotion?  \\
\textbf{Aoi Endo}: Support for management training and obtaining qualifications would be helpful.  \\
\textbf{System}: You mentioned that support for management training and obtaining qualifications would be helpful. Specifically, what qualifications would you like to obtain?  \\
\textbf{Aoi Endo}: I would like to obtain a nursing management qualification and take leadership training.  \\
\textbf{System}: What kind of support do you feel is necessary for obtaining the nursing management qualification?  \\
\textbf{Aoi Endo}: I need time off for studying and financial assistance.  \\
\textbf{System}: You need time off and financial assistance for obtaining qualifications. Regarding your interest in other hospitals, what kind of hospitals or environments are you specifically considering?  \\
\textbf{Aoi Endo}: A place where I can receive support for my family would be ideal.  \\
\textbf{System}: That's all for today! \\
\end{quote}

\subsection{An Example Dialogue using Proposed Method 2 with Aoi Endo (Originally in Japanese) }\label{ap:ex_dia_p-m2}
\begin{quote}
\textbf{System}: Have you been busy lately?  \\
\textbf{Aoi Endo}: Yes, I'm busy, but I feel it's rewarding.  \\
\textbf{System}: I see, it's wonderful that you're feeling rewarded! By the way, can you tell me about your current job responsibilities?  \\
\textbf{Aoi Endo}: I'm supporting the team as a deputy leader in the internal medicine department.  \\
\textbf{System}: As a deputy leader, can you tell me specifically what kind of tasks you're responsible for?  \\
\textbf{Aoi Endo}: I'm in charge of coordinating the team, training new staff, and creating care plans for patients.  \\
\textbf{System}: I see, you're handling a variety of roles. Regarding leadership skills, what specific skills do you want to improve?  \\
\textbf{Aoi Endo}: I'd especially like to strengthen my communication skills as part of my leadership skills.  \\
\textbf{System}: Can you tell me about your specific goals for the next year?  \\
\textbf{Aoi Endo}: Next year, I want to take the next step towards a nursing management position.  \\
\textbf{System}: As you're aiming to step into a nursing management position, what preparations are you currently making?  \\
\textbf{Aoi Endo}: I'm participating in management training and learning about leadership.  \\
\textbf{System}: You're participating in management training. Can you tell me more about the specific content of that training?  \\
\textbf{Aoi Endo}: In the training, I'm learning skills related to management and problem-solving.  \\
\textbf{System}: What kind of results or changes have you noticed since taking that training?  \\
\textbf{Aoi Endo}: My management skills have improved, and it's become easier to coordinate the team.  \\
\textbf{System}: What kind of results or changes are you hoping to achieve through this training?  \\
\textbf{Aoi Endo}: I'm aiming to improve the overall performance of the team and ensure smooth communication.  \\
\textbf{System}: Regarding the improvement of overall team performance, can you tell me more specifically what improvements you're aiming for?  \\
\textbf{Aoi Endo}: I'd like to improve the quality of training for new staff and streamline patient care.  \\
\textbf{System}: Regarding improving the quality of training for new staff, what specific improvements are you aiming for?  \\
\textbf{Aoi Endo}: I want to enhance practical guidance so that new staff can quickly adapt to the workplace.  \\
\textbf{System}: That's all for today! \\
\end{quote}

\section{An Example of Persona (Aoi Takahashi) and Dialogues }
\subsection{An Example of Persona Setting: Aoi Takahashi (Originally in Japanese)}\label{ap:persona_example_takahashi}
Table~\ref{tab:persona_example_takahashi} shows the persona setting of Aoi Takahashi.
\begin{table}[tbp] 
\centering 
\begin{tabular}{|>{\centering\arraybackslash}p{2cm}|p{4cm}|} \hline 
\textbf{Attribute} & \textbf{Details} \\ \hline 
\small \textbf{Name} & Aoi Takahashi \\ \hline 
\small \textbf{Age} & 32 years old \\ \hline 
\small \textbf{Hometown} & Osaka Prefecture \\ \hline 
\small \textbf{Gender} & Female \\ \hline 
\small \textbf{Personality} & Extroverted \\ \hline 
\small \textbf{Past Career} & A nurse working at a university hospital. Has been a nurse for 10 years, with experience in pediatrics for five years. \\ \hline 
\small \textbf{Current Career} & Currently works in orthopedics and considers herself a skilled nurse. \\ \hline 
\small \textbf{Future Aspirations} & Aims to balance family and work, and is considering becoming a generalist. \\ \hline 
\small \textbf{Thoughts} & Feels confident in her job. She is discussing work-life balance with her husband as her son will start elementary school next year. \\ \hline 
\small \textbf{Other Details} & Her son, who will start elementary school next year, wants to attend a soccer school after school. She lives with her husband and son. They often take walks in the local park on weekends. Her favorite TV show is "World Heritage Sites." \\ \hline 
\small \textbf{Check Items} & Intentions toward becoming a generalist, Concerns about work-life balance. \\ \hline 
\end{tabular} 
\caption{Persona: Aoi Takahashi.} 
\label{tab:persona_example_takahashi}
\end{table}

\subsection{An Example Dialogue using Baseline Method with Aoi Takahashi (Originally in Japanese) }
\begin{quote}
\textbf{System}: Have you been busy lately? \\
\textbf{Aoi}: I'm getting used to work, but I'm worried about my son starting elementary school. \\
\textbf{System}: I see, so your son is starting elementary school. What are you worried about? \\
\textbf{Aoi}: I'm anxious about balancing work and raising my child. \\
\textbf{System}: I see, that must be tough. Specifically, what are you concerned about? \\
\textbf{Aoi}: I'm worried about how my son will spend his time after school and getting called in suddenly. \\
\textbf{System}: That is concerning. Do you have anyone to support you? \\
\textbf{Aoi}: My husband is supportive, but I still want to be the main one taking care of him. \\
\textbf{System}: I see, family support is important. What kind of support do you think would help with your son's after-school time? \\
\textbf{Aoi}: My son wants to join a soccer school, and I'm worried about the pick-up and drop-off. \\
\textbf{System}: What kind of support do you think would help with the soccer school pick-up and drop-off? \\
\textbf{Aoi}: It would help if my husband and I could take turns with the pick-up and drop-off. What do you think? \\
\textbf{System}: That's a great idea! It might be a good idea to discuss it with your husband. Is there anything else you need support with? \\
\textbf{Aoi}: I want support for when I'm suddenly called in. \\
\textbf{System}: What kind of support do you think would help for sudden call-ins? \\
\textbf{Aoi}: Flexible shift arrangements at work would help. \\
\textbf{System}: You mentioned flexible shifts at work would help, but what kind of shift adjustments do you think would be ideal? \\
\textbf{Aoi}: A flexible shift that allows me to respond to sudden call-ins would be ideal. \\
\textbf{System}: Specifically, what kind of shift adjustments would be ideal? For example, is there a particular time of day or day of the week when you'd like more flexibility? \\
\textbf{Aoi}: It would help if I had flexibility in the afternoons or evenings on weekdays. \\
\textbf{System}: That's it for today! \\
\end{quote}
\subsection{An Example Dialogue using Proposed Method 1 with Aoi Takahashi (Originally in Japanese) }
\begin{quote}
\textbf{System}: Have you been busy lately? \\
\textbf{Aoi}: I'm getting used to work, but I'm worried about my son. \\
\textbf{System}: I see, that sounds tough. What are you worried about with your son, specifically? \\
\textbf{Aoi}: My son wants to go to soccer school, and it's hard to adjust my time. \\
\textbf{System}: That does sound difficult. Do you think you can adjust your shifts? Is there anything I can help with? \\
\textbf{Aoi}: Adjusting my shifts might be hard. I'll try discussing it with my husband. \\
\textbf{System}: That sounds like a good idea. I hope you and your husband can work something out. \\
\textbf{Aoi}: Thanks, I'll try that. I also need to keep up with work. \\
\textbf{System}: Work must be tough too. Could you tell me more about your job responsibilities? \\
\textbf{Aoi}: I handle general nursing duties in orthopedics. \\
\textbf{System}: Orthopedic nursing sounds tough. How satisfied are you with your job? \\
\textbf{Aoi}: I find my job fulfilling, but I get tired sometimes. \\
\textbf{System}: It must be exhausting at times. Do you have any particular department you'd like to be assigned to in the future? \\
\textbf{Aoi}: I'm aiming to be a generalist, so I'm not particular about a specific department. \\
\textbf{System}: I see, you're aiming to become a generalist. Is there any other support or concerns you have about your family? \\
\textbf{Aoi}: For now, everything is fine, but I'll consult you again if I need to. \\
\textbf{System}: Got it, feel free to ask anytime. By the way, could you tell me more specifically about your shift preferences? \\
\textbf{Aoi}: Which is easier to adjust, early or late shifts? \\
\textbf{System}: Could you tell me specifically what kind of shift adjustments you'd like? \\
\textbf{Aoi}: Early shifts might be easier to adjust because I have to pick up and drop off my son. \\
\textbf{System}: That's it for today! \\
\end{quote}

\subsection{An Example Dialogue using Proposed Method 2 with Aoi Takahashi (Originally in Japanese) }
\begin{quote}
\textbf{System}: Have you been busy lately? \\
\textbf{Aoi}: I'm getting used to work, but things have been hectic with my son. \\
\textbf{System}: I see, that sounds tough. Is there anything I can help with regarding your son? \\
\textbf{Aoi}: Thanks, but we're planning to manage it as a family. \\
\textbf{System}: Got it, don't push yourself too hard. Is there anything work-related you're struggling with? \\
\textbf{Aoi}: Balancing work and raising my child is difficult, and I'm not sure what to do. \\
\textbf{System}: I see, balancing both really is tough. Specifically, what part of it is difficult? \\
\textbf{Aoi}: My work shifts and my son's soccer school schedule overlap. \\
\textbf{System}: If you have any shift preferences, could you let me know? For example, if you need specific days or times off? \\
\textbf{Aoi}: I would like to have afternoons off on weekends, so I can accompany my son to soccer. \\
\textbf{System}: I see, so you'd like to have weekends off in the afternoon. Can you tell me more about your hopes for family time? \\
\textbf{Aoi}: I want to prioritize family time, so I'd like to spend as much time as possible with them on weekends. \\
\textbf{System}: Got it, it's wonderful to prioritize family time. By the way, could you tell me about your current job responsibilities? \\
\textbf{Aoi}: I work in orthopedics, assisting with surgeries and caring for patients. \\
\textbf{System}: Are you satisfied with your job in orthopedics, or is there something you'd like to change? \\
\textbf{Aoi}: I'm satisfied with my job, but I want to keep growing. \\
\textbf{System}: What specific goals do you have for growth? \\
\textbf{Aoi}: I'm aiming to be a generalist, so I want to gain broad knowledge and experience. \\
\textbf{System}: What areas or skills are you particularly interested in learning to gain broader experience? \\
\textbf{Aoi}: I want to learn more about emergency medicine and home healthcare. \\
\textbf{System}: That's it for today! \\
\end{quote}
\end{document}